\documentclass{article}
\usepackage{spconf,amsmath,graphicx}

\usepackage[english]{babel}
\usepackage[T1]{fontenc}  
\usepackage[latin1]{inputenc} 
\usepackage{url}
\usepackage{graphicx}
\usepackage{subfigure}
\usepackage{xcolor}
\usepackage{amsmath,amsfonts,amssymb}
\usepackage{booktabs}
\usepackage{epsfig}
\usepackage{pstricks}
\usepackage{pst-node}
\usepackage{pst-grad}
\usepackage{ifthen}
\usepackage{algorithm}
\usepackage{algorithmic}	
\usepackage{setspace}           

\usepackage{epstopdf}	
\usepackage{enumerate} 
\usepackage{hyperref}
\usepackage{multicol} 
\usepackage{enumitem} 
\usepackage{cite} 
\usepackage{appendix}

\allowdisplaybreaks

\newcommand{\mypar}[1]{\bigskip\noindent {\bf #1.}}

\definecolor{red}{RGB}{153,0,0}

\title{Coupled Dictionary Learning for Multi-contrast MRI Reconstruction}

\name{
	Pingfan Song$^{\star}$ \quad
	Lior Weizman$^{\sharp}$ \quad
	Jo\~ao F.\ C.\ Mota$^{\dagger}$ \quad  
	Yonina C. Eldar$^{\sharp}$ \quad
	Miguel R.\ D.\ Rodrigues$^{\star}$  
	\thanks{
		This work was supported by the Royal Society International Exchange Scheme IE160348, by the European Union's Horizon 2020 grant ERC-BNYQ, by the Israel Science Foundation grant no. 335/14, by ICore: the Israeli Excellence Center 'Circle of Light', by the Ministry of Science and Technology, Israel, by UCL Overseas Research Scholarship (UCL-ORS) and by China Scholarship Council (CSC).
	}
}
\address{				
	$^{\star}$ Department of Electronic and Electrical Engineering, University College London, UK \\
	$^{\sharp}$ Department of Electrical Engineering, Technion -- Israel Institute of Technology, Israel \\
	$^{\dagger}$ School of Engineering and Physical Sciences, Heriot-Watt University, UK
}

\begin{document}
\ninept 
\maketitle

\begin{abstract}
	Medical imaging tasks often involve multiple contrasts, such as T1- and T2-weighted magnetic resonance imaging (MRI) data. These contrasts capture information associated with the same underlying anatomy and thus exhibit similarities. In this paper, we propose a Coupled Dictionary Learning based multi-contrast MRI reconstruction (CDLMRI) approach to leverage an available guidance contrast to restore the target contrast. Our approach consists of three stages: coupled dictionary learning, coupled sparse denoising, and $k$-space consistency enforcing. The first stage learns a group of dictionaries that capture correlations among multiple contrasts. By capitalizing on the learned adaptive dictionaries, the second stage performs joint sparse coding to denoise the corrupted target image with the aid of a guidance contrast. The third stage enforces consistency between the denoised image and the measurements in the $k$-space domain. Numerical experiments on the retrospective under-sampling of clinical MR images demonstrate that incorporating additional guidance contrast via our design improves MRI reconstruction, compared to state-of-the-art approaches.
\end{abstract}

\section{Introduction}
\label{sec:Introduction}

\vspace{-0.2cm}

Magnetic Resonance Imaging (MRI) is a noninvasive and non-ionizing medical imaging technique that has been widely used for medical diagnosis, clinical analysis, and staging of disease. Owing to its versatility, different MRI pulse sequences produce images with different contrasts, such as Fluid-attenuated inversion recovery (FLAIR), T1-weighted, and T2-weighted. Each contrast images different physical properties of the tissues examined~\cite{liang2000principles, mcrobbie2007mri}. 
The acquisition time of conventional brain multi-contrast MRI is at least 30 minutes, which can lead to discomfort in some patients and requires sedation for pediatric patients. Different image constrasts, however, are highly correlated, because they image the same underlying anatomy~\cite{ehrhardt2016multicontrast}. Such correlation can potentially be used to shorten acquisition time by partial acquisition of the target contrast, followed by reconstruction that takes into account other fully-sampled contrasts as guidance/reference.

MRI reconstruction from under-sampled measurements has been thoroughly investigated in the case of single contrast acquisition. The pioneering framework proposed by Lustig et al.~\cite{Lustig07-SparseMRI}, motivated by compressive sensing theory~\cite{donoho2006compressed,candes2006robust,eldar2012compressed,eldar2015sampling}, uses the fact that MR images are sampled in the spatial frequency domain (a.k.a. $k$-space) and can be represented as a sparse combination of fixed, predefined bases, for example, wavelets. Building on this work, Ravishankar et al.~\cite{ravishankar2011mr} proposed a dictionary learning based MRI reconstruction approach, DLMRI, which is based on the fact that patches of MR images can be sparsely represented with respect to a set of adaptive learned bases~\cite{olshausen1996emergence,aharon2006img}. Those adaptive bases contribute to improved performance of DLMRI over SparseMRI~\cite{Lustig07-SparseMRI}. Both SparseMRI and DLMRI consider a single contrast.

The reconstruction of an MRI target contrast based on the availability of other contrasts has also been recently investigated in some works~\cite{weizman2016reference,weizman2015compressed,ehrhardt2016multicontrast,ehrhardt2015joint,huang2014fast,bilgic2011multi,qu2014magnetic,chatnuntawech2016vectorial}. For example, for the case where one contrast (the "target contrast") is under-sampled and the other is fully sampled and serves as guidance/reference, Weizman et al.~\cite{weizman2016reference,weizman2015compressed} proposed reference-based MRI, that exploits gray level similarity between T2-weighted and FLAIR, to reconstruct the target contrast (FLAIR) given the guidance contrast (T2-weighted). Their approach is specific to contrasts with gray level similarity, not to structural similarity which is our focus here. Ehrhardt et al.~\cite{ehrhardt2016multicontrast,ehrhardt2015joint} propose an approach, STVMRI, that exploits structure-guided total variation to integrate the location and direction priors from a guidance contrast into the reconstruction of the target contrast.

Inspired by DLMRI~\cite{ravishankar2011mr}, we propose a coupled dictionary learning approach for multi-contrast MRI reconstruction, referred to as CDLMRI, for contrasts that exhibit structural similarity (e.g. T1-weighted and T2-weighted), a scenario more general than the one addressed in~\cite{weizman2016reference}. Specifically, our approach cycles between three stages: coupled dictionary learning, coupled sparse denoising and $k$-space consistency enforcing. 
The first stage learns a group of dictionaries that capture inherent structural similarity on textures, edges, boundaries, or other salient features across multiple contrasts. The second stage performs joint sparse coding using the learned adaptive dictionaries to denoise the corrupted target image with the aid of a guidance contrast. The third stage enforces consistency between the denoised image and the measurements in the $k$-space domain.
%
%
Our approach shows significant advantages over the competing methods, DLMRI~\cite{ravishankar2011mr} and STVMRI~\cite{ehrhardt2016multicontrast} both in visual quality, and in peak signal-to-noise ratio (PSNR).

\vspace{-0.3cm}

\section{Problem Formulation}
\label{sec:Problem}

\vspace{-0.2cm}

	We denote by $\mathbf{x}^{(1)} \in \mathcal{C}^N $ the vectorized 2D MR imaging contrast of size $\sqrt{N} \times \sqrt{N}$ to be reconstructed. The vector $\mathbf{y}^{(1)} \in \mathcal{C}^{m}$ denotes the under-sampled $k$-space measurements related to $\mathbf{x}^{(1)}$, and the matrix $\mathbf{F}_{u1} \in \mathcal{C}^{m \times N}$ denotes the corresponding under-sampled Fourier transform matrix. In addition, we assume that a fully-sampled guidance MR imaging contrast $\mathbf{x}^{(2)}  \in \mathcal{C}^N $ is available. Our goal is to reconstruct $\mathbf{x}^{(1)}$ from its $k$-space samples $\mathbf{y}^{(1)}$ under the aid of the guidance image $\mathbf{x}^{(2)}$.
%


\vspace{-0.3cm}

\subsection{Data Model for Multi-contrast MRI Data}

\vspace{-0.2cm}

As we would like to utilize the structural similarity between $\mathbf{x}^{(1)}$ and $\mathbf{x}^{(2)}$, we first propose a data model that captures this similarity. Our data model works with image patches, instead of the entire image level, because a patch-based model is able to capture local image features effectively, as shown in other applications such as image denoising, super-resolution, inpainting, deblurring and demosaicing~\cite{elad2006image,chatterjee2012patch,mairal2014sparse,mairal2009non,timofte2013anchored,timofte2014a+,yang2010image}.

Let $\mathbf{x}_{ij}^{(1)} \in \mathcal{C}^n $ and $\mathbf{x}_{ij}^{(2)} \in \mathcal{C}^n$ denote the vector representations of image patch pairs of size $\sqrt{n} \times \sqrt{n}$ extracted from the image $\mathbf{x}^{(1)}$ and $\mathbf{x}^{(2)}$, respectively, where the tuple $(i, j)$ denotes the coordinates of the top-left corner of the patches within the images. Formally, we write $\mathbf{x}_{ij}^{(1)} = \mathbf{R}_{ij} \mathbf{x}^{(1)}$ (resp. $\mathbf{x}_{ij}^{(2)} = \mathbf{R}_{ij} \mathbf{x}^{(2)}$), where the matrix $\mathbf{R}_{ij} \in \mathcal{C}^{n \times N}$ represents the operator that extracts patch $\mathbf{x}_{ij}^{(1)}$ (resp. $\mathbf{x}_{ij}^{(2)}$) from $\mathbf{x}^{(1)}$ (resp. $\mathbf{x}^{(2)}$). 
%
In order to capture both the similarity and discrepancy between two different contrasts, we assume that each patch pair $(\mathbf{x}^{(1)}_{ij}, \mathbf{x}^{(2)}_{ij} )$ can be represented by a sum of two sparse representations: a common sparse component that is shared by both contrasts, and a unique sparse component for each contrast. In particular, we express the patch pair as follows:
\vspace{-0.2cm}
\begin{align} 
	\mathbf{x}_{ij}^{(1)} &= \boldsymbol{\Psi}_c \, \mathbf{z}_{ij} + \boldsymbol{\Psi} \, \mathbf{u}_{ij} \,,
	\label{Eq:SparseRepreMultiMRI1}
	\\
	\mathbf{x}_{ij}^{(2)} &= \boldsymbol{\Phi}_c \, \mathbf{z}_{ij} + \boldsymbol{\Phi} \, \mathbf{v}_{ij} \,,
	\label{Eq:SparseRepreMultiMRI2}
\end{align}
\vspace{-0.5cm}

\noindent where $\mathbf{z}_{ij} \in \mathbb{R}^{K}$ is the common sparse representation shared by both contrasts, $\mathbf{u}_{ij} \in \mathbb{R}^{K}$ is a sparse representation specific to contrast $\mathbf{x}^{(1)}$, and $\mathbf{v}_{ij} \in \mathbb{R}^{K}$ is a sparse representation specific to contrast $\mathbf{x}^{(2)}$; in addition, $\boldsymbol{\Psi}_{c} = [\boldsymbol{\psi}_{c1}, \cdots, \boldsymbol{\psi}_{cK}] \in \mathbb{R}^{n \times K}$ and $\boldsymbol{\Phi}_{c} = [\boldsymbol{\phi}_{c1}, \cdots, \boldsymbol{\phi}_{cK}] \in \mathbb{R}^{n \times K}$ are a pair of dictionaries associated with the common sparse representation $\mathbf{z}_{ij}$, whereas $\boldsymbol{\Psi} = [\boldsymbol{\psi}_{1}, \cdots, \boldsymbol{\psi}_{K}] \in \mathbb{R}^{n \times K}$ and $\boldsymbol{\Phi} = [\boldsymbol{\phi}_{1}, \cdots, \boldsymbol{\phi}_{K}] \in \mathbb{R}^{n \times K}$ are dictionaries associated with the sparse representations $\mathbf{u}_{ij}$ and $\mathbf{v}_{ij}$, respectively. Note that, this model can be generalized for $\boldsymbol{\Psi}$ and $\boldsymbol{\Phi}$ to have different number of atoms.



\vspace{-0.3cm}

\subsection{CDLMRI}
\label{ssec:CDLMRI}

\vspace{-0.2cm}

Our goal is to leverage the proposed data model~\eqref{Eq:SparseRepreMultiMRI1}-\eqref{Eq:SparseRepreMultiMRI2} in order to recover the target MRI contrast $\mathbf{x}^{(1)}$ given the guidance MRI contrast $\mathbf{x}^{(2)}$ and the $k$-space measurements $\mathbf{y}^{(1)}$. To this end, we propose Coupled Dictionary Learning for multi-contrast MRI reconstruction algorithm that attempts to solve the following optimization problem
%
\vspace{-0.2cm}
\begin{equation} \label{Eq:CDLMRI}
\small
\begin{array}{cl}
\underset{
	\begin{subarray}{c}
	\mathbf{x}^{(1)}, \mathbf{z}_{ij}, \mathbf{u}_{ij}, \mathbf{v}_{ij} \\
	\boldsymbol{\Phi}_c, \boldsymbol{\Phi},  \boldsymbol{\Psi}_c, \boldsymbol{\Psi}
	\end{subarray}
}{\text{minimize}}
& \!\!\!\!	
\sum\limits_{ij} 
\Big\{ \| \mathbf{R}_{ij} \mathbf{x}^{(1)} - ( \boldsymbol{\Psi}_c \mathbf{z}_{ij} + \boldsymbol{\Psi} \mathbf{u}_{ij} ) \|_2^2 
\\
& \!\!\!\! 
\quad
+ \| \mathbf{R}_{ij} \mathbf{x}^{(2)} - ( \boldsymbol{\Phi}_c \mathbf{z}_{ij} + \boldsymbol{\Phi} \mathbf{v}_{ij} ) \|_2^2 
\\
& \!\!\!\! 
\quad
+ \nu_1 \|\mathbf{F}_{u1} \mathbf{x}^{(1)} - \mathbf{y}^{(1)}  \|_2^2
\Big\}
\\
\text{subject to}
& \!\!\!\! 
\|\mathbf{z}_{ij} \|_0 \leq s_c, 
\|\mathbf{u}_{ij} \|_0 \leq s_1,
\|\mathbf{v}_{ij} \|_0 \leq s_2,
\, \forall i,j,
\\
& \!\!\!\! 
\left\| \begin{bmatrix}
	\boldsymbol{\psi}_{ck} \\ \boldsymbol{\phi}_{ck} 
\end{bmatrix} \right\|_2^2 \leq 1, 
\|\boldsymbol{\psi}_k \|_2^2 \leq 1, 
\|\boldsymbol{\phi}_k \|_2^2 \leq 1, 
\, \forall k.
\end{array}
\end{equation}
Note that the first two terms in the objective ensure that the image patches are consistent with their postulated model and the third term in the objective ensures that the target image is consistent with its $k$-space measurements; the parameter $\nu_1 \geq 0$ balances between model and measurements fidelity. Moreover, the first set of constraints induce sparsity for the vectors $\mathbf{z}_{ij},\mathbf{u}_{ij},\mathbf{v}_{ij}$ and the second set of contraints normalizes the atoms of the dictionaries in order to remove the scaling ambiguity and avoid trivial solutions. 

%

Problem~\eqref{Eq:CDLMRI} is highly nonconvex. Therefore, we attempt to solve this problem by alternating between three stages over a number of cycles: 1) Coupled dictionary learning, 2) Coupled sparse denoising, and 3) $k$-space consistency enforcing, as shown in  Algorithm~\ref{Alg:CDLMRI}. 


\begin{algorithm}[t] 
	\caption{CDLMRI Algorithm}
	\label{Alg:CDLMRI}
	\begin{algorithmic}[1]	
		\renewcommand{\algorithmicrequire}{\textbf{Input:}}
		\renewcommand{\algorithmicensure}{\textbf{Output:}}
		\REQUIRE 
		Under-sampled $k$-space measurements: $\mathbf{y}^{(1)} $; Guidance contrast: $\mathbf{x}^{(2)}$; Parameter: $\nu_1$; Sparse constraints: $s_c$, $s_1$, $s_2$; Number of iterations for CDL: $L$; Number of cycles for CDLMRI: $T$; Number of overlapping patches at each pixel: $\beta$.
		
		\ENSURE
		Estimated $\tilde{\mathbf{x}}^{(1)}$.
		
		\renewcommand{\algorithmicrequire}{\textbf{Initialization:}}
		\REQUIRE
		Initialize $\mathbf{x}^{(1)}$ as $\mathbf{x}^{(1)} = \mathbf{F}_{u1}^H \mathbf{y}^{(1)}$.

		\renewcommand{\algorithmicrequire}{\textbf{Optimization:}}
		\REQUIRE		
		\FOR{$t= 1, \cdots, T$} 
		\STATE
		\textbf{Coupled Dictionary Learning stage.}
		Estimate the coupled dictionaries $\boldsymbol{\Psi}_c$, $\boldsymbol{\Psi}$, $\boldsymbol{\Phi}_c$, $\boldsymbol{\Phi}$ using Algorithm~\ref{Alg:CoupledBCD}.
		
		\STATE
		\textbf{Coupled Sparse Denoising stage.}
		Estimate the sparse representations $\mathbf{z}_{ij}$ and $\mathbf{u}_{ij}$ by solving~\eqref{Eq:SC_Com} and \eqref{Eq:SC_Unique1} using the OMP algorithm. It also involves estimating the denoised patches $\hat{\mathbf{x}}^{(1)}_{ij} = \boldsymbol{\Psi}_c \mathbf{z}_{ij} + \boldsymbol{\Psi} \mathbf{u}_{ij} $ from $\mathbf{z}_{ij}$, $\mathbf{u}_{ij}$.
		
		\STATE
		\textbf{$k$-space Consistency Enforcing stage.}
		Estimate the target contrast $\tilde{\mathbf{x}}^{(1)}$ from estimated $k$-space samples $\tilde{\mathbf{y}}^{(1)}$ via~\eqref{Eq:CDLMRI_ReconUpdateEnd} and \eqref{Eq:CDLMRI_ReconUpdate9}.
		
		\ENDFOR
	\end{algorithmic}
\end{algorithm}

\vspace{-0.2cm}

\mypar{Stage 1) Coupled Dictionary Learning}
In the first cycle, $\mathbf{x}^{(1)}$ is initialized as $\mathbf{F}_{u1}^H \mathbf{y}^{(1)}$ (i.e. $\mathbf{x}^{(1)}$ is set to be equal to the inverse DFT of the zero-filled Fourier measurements). In the remaining cycles, in this stage, $\mathbf{x}^{(1)}$ will be the output of Stage 3) from the previous cycle. In particular, for fixed $\mathbf{x}^{(1)}$, we attempt to solve~\eqref{Eq:CDLMRI} via alternating minimization where in a first step we update $\mathbf{z}_{ij}, \mathbf{u}_{ij}, \mathbf{v}_{ij}$ for fixed $\boldsymbol{\Psi}_c, \boldsymbol{\Psi}, \boldsymbol{\Phi}_c, \boldsymbol{\Phi}$ and in a second step we update $\boldsymbol{\Psi}_c, \boldsymbol{\Psi}, \boldsymbol{\Phi}_c, \boldsymbol{\Phi}$ for fixed $\mathbf{z}_{ij}, \mathbf{u}_{ij}, \mathbf{v}_{ij}$. As shown in Algorithm~\ref{Alg:CoupledBCD}, the sparse coding step is addressed using the orthogonal matching pursuit (OMP) algorithm~\cite{tropp2007signal,mairal2014sparse} and dictionary update step is adapted from the Block Coordinate Descent~\cite{mairal2010online}. Note that, since a single image consists of large amount of patches, we only use a subset of the patches to constitute the training dataset $\mathbf{X}^{(1)}=[\cdots, \mathbf{x}_{ij}^{(1)}, \cdots]$ and $\mathbf{X}^{(2)}=[\cdots, \mathbf{x}_{ij}^{(2)}, \cdots]$ in Stage 1) to save training time.
\begin{algorithm*}[t] 
	\caption{Coupled Dictionary Learning algorithm}
	\label{Alg:CoupledBCD}
	\begin{multicols}{2}  
		\begin{algorithmic}[1]	
			\renewcommand{\algorithmicrequire}{\textbf{Input:}}
			\renewcommand{\algorithmicensure}{\textbf{Output:}}
			\REQUIRE 
			
			A subset of estimated target image patches: 
			$\mathbf{X}^{(1)}=[\cdots, \mathbf{x}_{ij}^{(1)}, \cdots]$; 
			The subset of the corresponding guidance image patches: 
			$\mathbf{X}^{(2)}=[\cdots, \mathbf{x}_{ij}^{(2)}, \cdots]$; 
			Sparse constraints: $s_c$, $s_1$, $s_2$; Number of iterations: $L$; \\
			
			\ENSURE 
			Coupled dictionaries: $\boldsymbol{\Psi}_{c}, \boldsymbol{\Phi}_{c},\boldsymbol{\Psi}, \boldsymbol{\Phi}$.
			
			\renewcommand{\algorithmicrequire}{\textbf{Initialization:}}
			\REQUIRE
			Initialize each dictionary with randomly selected patches of the corresponding contrast. Initialize all sparse representations with zeros.
			
			\renewcommand{\algorithmicrequire}{\textbf{Optimization:}}
			\REQUIRE		
			\FOR{$l= 1, \cdots, L$} 
			\STATE
			\textbf{a) Sparse Coding step.} 
			This step updates the sparse codes for fixed dictionaries.
			Note that steps \ref{Alg:OMPstart} - \ref{Alg:OMPend} apply the OMP algorithm~\cite{tropp2007signal} to estimate $\mathbf{Z} = [\cdots, \mathbf{z}_{ij}, \cdots]$ given data $\mathbf{X}^{(1)}$, $\mathbf{X}^{(2)}$ and dictionaries $\boldsymbol{\Psi}_c$, $\boldsymbol{\Phi}_c$.
			
			\STATE 
			\label{Alg:OMPstart}
			Initialize the active set $\Gamma=\emptyset$ and $\mathbf{z}_{ij} \leftarrow 0$.
			\WHILE{$|\Gamma| < s_c$ } 
			\STATE
			select a new coordinate $\hat{k}$ that leads to the smallest residual and, then update the active set and the solution $\mathbf{z}_{ij}$:
			\begin{align*}
			&
			(\hat{k},\hat{\boldsymbol{\alpha}}) \in 
			\underset{k \in \Gamma^c, \boldsymbol{\alpha} \in \mathbb{R}^{|\Gamma|+1}}{\arg \min } \;
			\left\| 
			\begin{bmatrix} 
			\mathbf{x}_{ij}^{(1)}  \\ 
			\mathbf{x}_{ij}^{(2)} 
			\end{bmatrix} 
			- 	
			\begin{bmatrix} 
			\boldsymbol{\Psi}_c \\ 
			\boldsymbol{\Phi}_c
			\end{bmatrix}_{\Gamma \cup \{k\}}
			\boldsymbol{\alpha}
			\right\|_2^2;
			\\
			&
			\Gamma \leftarrow \Gamma \cup \{\hat{k}\}; \quad
			\mathbf{z}_{ij_\Gamma} \leftarrow \hat{\boldsymbol{\alpha}}; \quad
			\mathbf{z}_{ij_{\Gamma^c}} \leftarrow 0;
			\end{align*}
			\ENDWHILE
			\label{Alg:OMPend}
			
			\STATE
			This OMP process is repeated to estimate $\mathbf{U} = [\cdots, \mathbf{u}_{ij}, \cdots]$ and $\mathbf{V} = [\cdots, \mathbf{v}_{ij}, \cdots]$ given $\mathbf{X}^{(1)} - \boldsymbol{\Psi}_c \mathbf{Z} = \boldsymbol{\Psi} \mathbf{U}$ and $\mathbf{X}^{(2)} - \boldsymbol{\Phi}_c \mathbf{Z} = \boldsymbol{\Phi} \mathbf{V}$, with sparsity constraints $s_1$ and $s_2$, respectively.
			
			
			
			\STATE
			\textbf{b) Dictionary Update step.} 
			This step updates the dictionaries for fixed sparse codes.
			
			\FOR{$k = 1, \cdots, K$}
			\STATE
			Update the $k$-th column of $\boldsymbol{\Psi}_{c}$ and $\boldsymbol{\Phi}_{c}$ as follows:
			\begin{align*}
			&
			\mathbf{d}_k  
			\leftarrow \frac{1}{\mathbf{z}^k {\mathbf{z}^k}^T}
			\left(
			\begin{bmatrix} 
			\mathbf{X}^{(1)} - \boldsymbol{\Psi} \mathbf{U} \\ 
			\mathbf{X}^{(2)} -  \boldsymbol{\Phi} \mathbf{V}
			\end{bmatrix} 
			- 	
			\begin{bmatrix} 
			\boldsymbol{\Psi}_c \\ 
			\boldsymbol{\Phi}_c
			\end{bmatrix} 
			\mathbf{Z} \right)
			{\mathbf{z}^k}^T + 
			\begin{bmatrix}
			\boldsymbol{\psi}_{ck}  \\
			\boldsymbol{\phi}_{ck}  \\
			\end{bmatrix}
			\\
			&
			\begin{bmatrix}
			\boldsymbol{\psi}_{ck}  \\
			\boldsymbol{\phi}_{ck}  \\
			\end{bmatrix}
			\leftarrow \frac{\mathbf{d}_k}{\max(\|\mathbf{d}_k \|_2, 1)} 
			\end{align*}
			where $\mathbf{z}^k$ denotes the $k$-th row of $\mathbf{Z}$.
			\ENDFOR
			
			\FOR{$k = 1, \cdots, K$}
			\STATE
			Update the $k$-th column of $\boldsymbol{\Psi}$ and $\boldsymbol{\Phi}$ as follows:
			\begin{align*}
			\boldsymbol{\psi}_k &\leftarrow \frac{1}{\mathbf{u}^k {\mathbf{u}^k}^T} 
			(\mathbf{X}^{(1)} - \boldsymbol{\Psi}_c \mathbf{Z} - \boldsymbol{\Psi} \mathbf{U}) {\mathbf{u}^k}^T + \boldsymbol{\psi}_k 
			\\
			\boldsymbol{\phi}_k &\leftarrow \frac{1}{\mathbf{v}^k {\mathbf{v}^k}^T} 
			(\mathbf{X}^{(2)} - \boldsymbol{\Phi}_c \mathbf{Z} - \boldsymbol{\Phi} \mathbf{V}) {\mathbf{v}^k}^T + \boldsymbol{\phi}_k 
			\\
			\boldsymbol{\phi}_k
			&\leftarrow \frac{\boldsymbol{\phi}_k}{\max(\|\boldsymbol{\phi}_k \|_2, 1)} 
			\;;
			\boldsymbol{\psi}_k
			\leftarrow \frac{\boldsymbol{\psi}_k}{\max(\|\boldsymbol{\psi}_k \|_2, 1)} 
			\end{align*}
			where $\mathbf{u}^k$ (resp. $\mathbf{v}^k$) denotes the $k$-th row of $\mathbf{U}$ (resp. $\mathbf{V}$).
			
			\ENDFOR
			\ENDFOR
			
			%
		\end{algorithmic}
	\end{multicols}
\end{algorithm*}	

\vspace{-0.2cm}

\mypar{Stage 2) Coupled Sparse Denoising}
As the sparse representations computed in Stage 1) are associated only with a subset of the collection of image patches, it is necessary to perform one additional sparse denoising stage. In addition, we also introduce linearly decreasing error thresholds to fine tune the sparse representations, since this operation has been shown experimentally to improve the de-aliasing and denoising performance. 
In particular, we give priority to perform the following optimization in order to determine an approximation to the common sparse representations associated with the various image patches:
%
%
%
\begin{equation} \label{Eq:SC_Com}
\small
\begin{array}{cl}
\underset{
	\begin{subarray}{c}
	\mathbf{z}_{ij}
	\end{subarray}
}{\text{min}}
& \!\!\!\!			
\text{max}\big\{
\| \mathbf{R}_{ij} \mathbf{x}^{(1)} - \boldsymbol{\Psi}_c \mathbf{z}_{ij} \|_2^2 
+
\| \mathbf{R}_{ij} \mathbf{x}^{(2)} - \boldsymbol{\Phi}_c \mathbf{z}_{ij}  \|_2^2 
- \epsilon_c, 0 \big\}
\\
\text{s.t.}
& \!\!\!\!
\|\mathbf{z}_{ij} \|_0 \leq s_c. 
\end{array}
\end{equation}
\noindent
We then perform the following optimization in order to determine an approximation to the unique sparse representations associated with the various target image patches:
%
\begin{equation} \label{Eq:SC_Unique1}
\small
\begin{array}{cl}
\underset{
	\begin{subarray}{c}
	\mathbf{u}_{ij}
	\end{subarray}
}{\text{min}}
&			
\text{max}\big\{
\| \mathbf{R}_{ij} \mathbf{x}^{(1)} - \boldsymbol{\Psi}_c \mathbf{z}_{ij} - \boldsymbol{\Psi} \mathbf{u}_{ij} \|_2^2 
- \epsilon_1, 0 \big\}
\\
\text{s.t.}
&
\|\mathbf{u}_{ij} \|_0 \leq s_1. 
\end{array}
\end{equation}
\noindent
Here $\epsilon_c$ and $\epsilon_1$ denote the expected error thresholds which are used, together with $s_c$ and $s_1$, in OMP as the stopping criteria. The above formulations imply that once the objective value for $(i,j)$-th patch decreases below the expected error threshold, there is no need to find a better point and so we terminate the OMP loop. Otherwise, the OMP program keeps iterating until $s_c$ (resp. $s_1$) non-zero values are retrieved for $\mathbf{z}_{ij}$ (resp. $\mathbf{u}_{ij}$).
In addition, as the quality of the target image improves along the entire cycles, we decrease the thresholds $\epsilon_c$ and $\epsilon_1$ linearly at each cycle. This strategy significantly accelerates the convergence speed, as well as allows us to dynamically control the real sparsity of each patch more effectively.

Given the sparse representations $\mathbf{z}_{ij}$ and $\mathbf{u}_{ij}$, we obtain each denoised patch as $\hat{\mathbf{x}}^{(1)}_{ij} = \boldsymbol{\Psi}_c \mathbf{z}_{ij} + \boldsymbol{\Psi} \mathbf{u}_{ij} $, where $\boldsymbol{\Psi}_c$ and $\boldsymbol{\Psi}$ are learned dictionaries in the Stage 1).

%
 

\vspace{-0.2cm}

\mypar{Stage 3) $k$-space Consistency Enforcing}
Finally, in this stage, we enforce the consistency between the denoised image and its measurements in the $k$-space domain, similar to DLMRI~\cite{ravishankar2011mr}. In particular, given the estimated patches $\hat{\mathbf{x}}^{(1)}_{ij}$ from Stage 2), this step is formulated as a least squares problem:
\begin{equation} \label{Eq:CDLMRI_ReconUpdate1}
\begin{array}{cl}
\underset{
	\begin{subarray}{c}
	\mathbf{x}^{(1)}
	\end{subarray}
}{\text{min}}
& 
\!\!\!\!			
\sum\limits_{ij} \left\| \mathbf{R}_{ij} \mathbf{x}^{(1)} - \hat{\mathbf{x}}^{(1)}_{ij} \right\|_2^2
+ \nu_1 \left \|\mathbf{F}_{u1} \mathbf{x}^{(1)} - \mathbf{y}^{(1)}  \right \|_2^2
\end{array} \,,
\end{equation}
By assuming that patches wrap around at image boundaries (which implies that the number of overlapping patches occurring at each pixel is equal), we immediately obtain the solution:
\begin{equation} \label{Eq:CDLMRI_ReconUpdateEnd}
\vspace{-0.2cm}
\tilde{\mathbf{x}}^{(1)}=\mathbf{F}^H \tilde{\mathbf{y}}^{(1)}
\end{equation}
where $\mathbf{F}^H$ denotes the conjugate of the Fourier transform matrix; $\tilde{\mathbf{y}}^{(1)}$ denotes the estimated $k$-space samples and can be expressed as follows:
\begin{equation} \label{Eq:CDLMRI_ReconUpdate9}
\tilde{\mathbf{y}}^{(1)}_{pq} = 
\left\{
\begin{aligned}
& \left( \mathbf{F} \hat{\mathbf{x}}^{(1)} \right)_{pq}, 	& \!\!\!\! (p, q) \notin \Omega^{(1)}
\\
&  \frac{ 1 }{1 + \tilde{\nu}_1} \left( \mathbf{F} \hat{\mathbf{x}}^{(1)} + \tilde{\nu}_1 \mathbf{F} \mathbf{F}_{u1}^H \mathbf{y}^{(1)} \right)_{pq}, 	& \!\!\!\! (p, q) \in \Omega^{(1)}
\end{aligned}
\right.
\end{equation}

\noindent
where $\tilde{\nu}_1 = \nu_1 / \beta$, $\hat{\mathbf{x}}^{(1)}=\frac{1}{\beta} \sum_{ij} \mathbf{R}_{ij}^H \hat{\mathbf{x}}^{(1)}_{ij} $ represents the denoised image, and $\beta$ denotes the number of overlapping patches at the corresponding pixel location in $\mathbf{x}^{(1)}$.
We denote by $\Omega^{(1)}$ the subset of $k$-space that has been sampled and by $\tilde{\mathbf{y}}^{(1)}_{pq}$ the updated value at location ($p, q$) in the $k$-space. 

The overall process consisting of the three stages is repeated over a number of cycles, which has been summarized in Algorithm~\ref{Alg:CDLMRI}.

\vspace{-0.2cm}

\section{Experiments}
\label{sec:Experiments}

\begin{figure*}[th]
	\begin{multicols}{2}  
		\centering
		\begin{minipage}[b]{0.3\linewidth}
			\raggedright
			\includegraphics[height = 2.2cm]{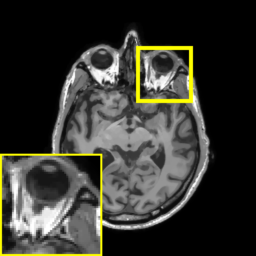}
			\scriptsize True T1
		\end{minipage} 
		\begin{minipage}[b]{0.3\linewidth}
			\raggedright
			\includegraphics[height = 2.2cm]{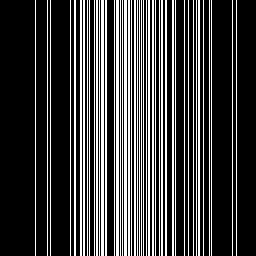}
			\scriptsize Mask
		\end{minipage} 
		\begin{minipage}[b]{0.3\linewidth}
			\raggedright
			\includegraphics[height = 2.2cm]{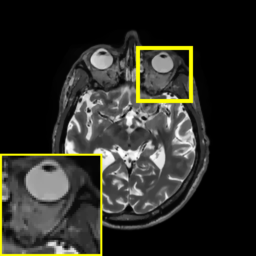}
			\scriptsize Guidance T2
		\end{minipage} 
		\\
		\vspace{0.2cm}
		\begin{minipage}[b]{0.3\linewidth}
			\raggedright
			\includegraphics[height = 2.2cm]{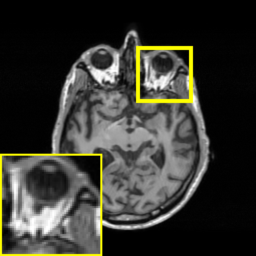}
		\end{minipage} 
		\begin{minipage}[b]{0.3\linewidth}
			\raggedright
			\includegraphics[height = 2.2cm]{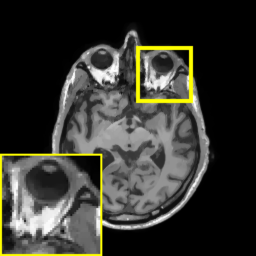}
		\end{minipage} 
		\begin{minipage}[b]{0.3\linewidth}
			\raggedright
			\includegraphics[height = 2.2cm]{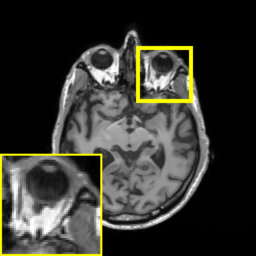}
		\end{minipage} 
		\\
		\begin{minipage}[b]{0.3\linewidth}
			\raggedright
			\includegraphics[height = 2.2cm]{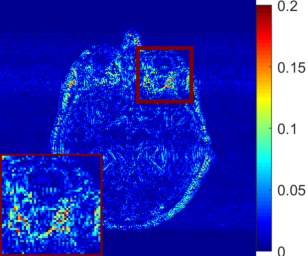}
		\end{minipage} 
		\begin{minipage}[b]{0.3\linewidth}
			\raggedright
			\includegraphics[height = 2.2cm]{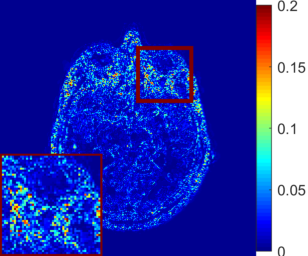}
		\end{minipage} 
		\begin{minipage}[b]{0.3\linewidth}
			\raggedright
			\includegraphics[height = 2.2cm]{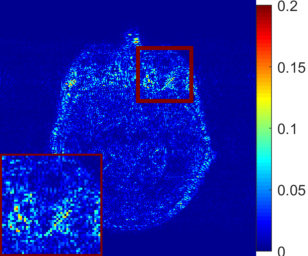}
		\end{minipage} 
		\\
		\begin{minipage}[b]{0.3\linewidth}
			\raggedright \scriptsize DLMRI, \\PSNR = 33.4dB. 
		\end{minipage} 
		\begin{minipage}[b]{0.3\linewidth}
			\raggedright \scriptsize STVMRI, \\PSNR = 33.7dB. 
		\end{minipage} 
		\begin{minipage}[b]{0.3\linewidth}
			\raggedright \scriptsize CDLMRI, \\PSNR = 36.1dB. 
		\end{minipage} 
		\\
		\begin{minipage}[t]{\linewidth}
			\centering \footnotesize (a) 4 fold 1D random under-sampling. \\
			\phantom{{\scriptsize leave some space.}}
		\end{minipage} 
		\\
		\begin{minipage}[b]{0.3\linewidth}
			\raggedright
			\includegraphics[height = 2.2cm]{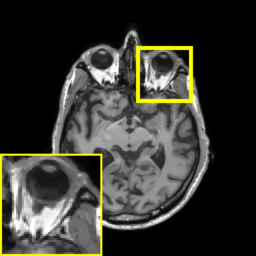}
			\scriptsize True T1
		\end{minipage} 
		\begin{minipage}[b]{0.3\linewidth}
			\raggedright
			\includegraphics[height = 2.2cm]{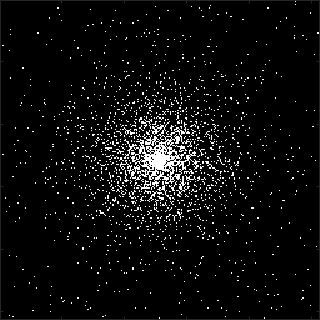}
			\scriptsize Mask
		\end{minipage} 
		\begin{minipage}[b]{0.3\linewidth}
			\raggedright
			\includegraphics[height = 2.2cm]{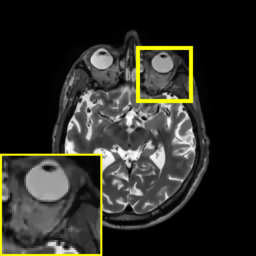}
			\scriptsize Guidance T2
		\end{minipage} 
		\\
		\vspace{0.2cm}
		\begin{minipage}[b]{0.3\linewidth}
			\raggedright
			\includegraphics[height = 2.2cm]{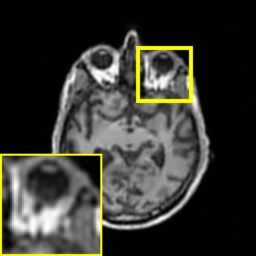}
		\end{minipage} 
		\begin{minipage}[b]{0.3\linewidth}
			\raggedright
			\includegraphics[height = 2.2cm]{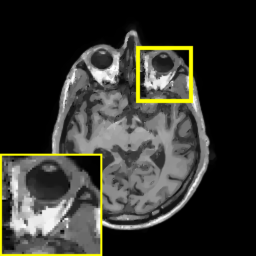}
		\end{minipage} 
		\begin{minipage}[b]{0.3\linewidth}
			\raggedright
			\includegraphics[height = 2.2cm]{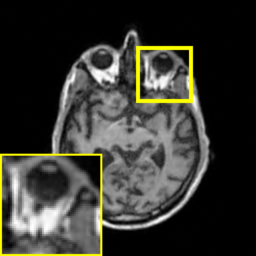}
		\end{minipage} 
		\\
		\begin{minipage}[b]{0.3\linewidth}
			\raggedright
			\includegraphics[height = 2.2cm]{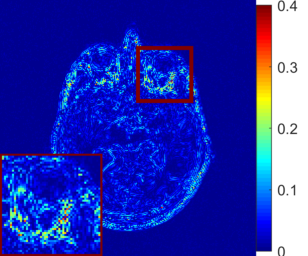}
		\end{minipage} 
		\begin{minipage}[b]{0.3\linewidth}
			\raggedright
			\includegraphics[height = 2.2cm]{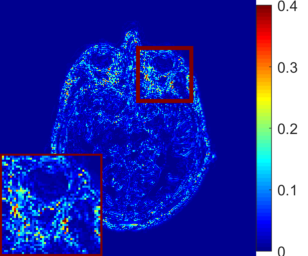}
		\end{minipage} 
		\begin{minipage}[b]{0.3\linewidth}
			\raggedright
			\includegraphics[height = 2.2cm]{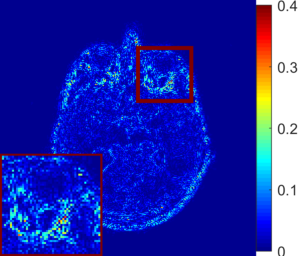}
		\end{minipage} 
		\\
		\begin{minipage}[b]{0.3\linewidth}
			\raggedright \scriptsize DLMRI, \\PSNR = 28.9dB. 
		\end{minipage} 
		\begin{minipage}[b]{0.3\linewidth}
			\raggedright \scriptsize STVMRI, \\PSNR = 29.0dB. 
		\end{minipage} 
		\begin{minipage}[b]{0.3\linewidth}
			\raggedright \scriptsize CDLMRI, \\PSNR = 30.2dB. 
		\end{minipage} 
		\begin{minipage}[t]{\linewidth}
			\centering \footnotesize (b) 20 fold 2D random under-sampling. 
		\end{minipage} 
	\end{multicols}	
	
	\vspace{-0.8cm}
	
	\caption{Reconstruction from Cartesian under-sampling. In each sub-figure, the first row shows the groundtruth T1-weighted, sampling mask and guidance modality T2-weighted. The second and third rows show the reconstructed images and the corresponding residual error from DLMRI~\cite{ravishankar2011mr}, STVMRI~\cite{ehrhardt2016multicontrast}, and the proposed approaches. It can be seen that the proposed approach reliably reconstructs fine details and substantially suppresses aliasing, noise and artifacts, leading to the smallest residual error.}
	\label{Fig:TestIm4}
\end{figure*}

\vspace{-0.2cm}

In this section, we conduct some practical experiments to evaluate the performance of the proposed algorithm. Similar to previous approaches\cite{ma2008efficient,yang2010fast,trzasko2009highly,ravishankar2011mr,ehrhardt2016multicontrast,weizman2016reference}, the data acquisition was simulated by retrospectively under-sampling the 2D discrete Fourier transform of clinical magnitude MR images. The sampling masks include Cartesian 1D and 2D random sampling.
%
%
%
%
%
We compare the proposed approach with DLMRI~\cite{ravishankar2011mr} to show the benefits of integrating guidance information into the MRI reconstruction task. DLMRI~\cite{ravishankar2011mr} is also based on dictionary learning techniques, but it does not use a guidance contrast to aid the reconstruction of the target contrast. We also compare with SVTMRI~\cite{ehrhardt2016multicontrast} which is also based on the use of a guidance contrast to aid the reconstruction of the target one.


In the experiments, we set $\sqrt{N} \times \sqrt{N} = 256 \times 256$, $\sqrt{n} \times \sqrt{n} = 8 \times 8$, $K=512$, $L=50$, $T=60$, $s_c=6, s_1=s_2=2$, $\epsilon_c=0.1\downarrow0.005$ (meaning: $\epsilon_c$ is set to 0.1 in the beginning and linearly decreases to 0.005 along the cycles.), $\epsilon_1=0.09\downarrow0.004$, $\beta=64$, $\nu_1 = \infty$ (for noise-free situation), and undersampling factor = 4 fold and 20 fold for the Cartesian 1D and 2D random sampling, respectively. 

The visual performance is shown in Fig.~\ref{Fig:TestIm4}. It can be seen that the reconstructed image and corresponding residual from DLMRI~\cite{ravishankar2011mr} introduce noticeable aliasing, noise and blurred areas. In comparison, the edges and outlines in the reconstructed image from STVMRI~\cite{ehrhardt2016multicontrast} are very sharp, thereby more visually appealing in some high-frequency areas. However, notice that some areas in the results of STVMRI~\cite{ehrhardt2016multicontrast} have been over-sharpened, thus introducing nonnegligible artifacts. In contrast, our approach substantially attenuates aliasing and noise and, at the same time, reliably restores fine details and suppresses artifacts, leading to a more comprehensive and interpretable reconstruction. The performance improvement is also demonstrated by the PSNR values.

\begin{figure}[h]
	\centering
	\begin{minipage}[b]{0.9\linewidth}
		\centering
		\includegraphics[width = 8cm]{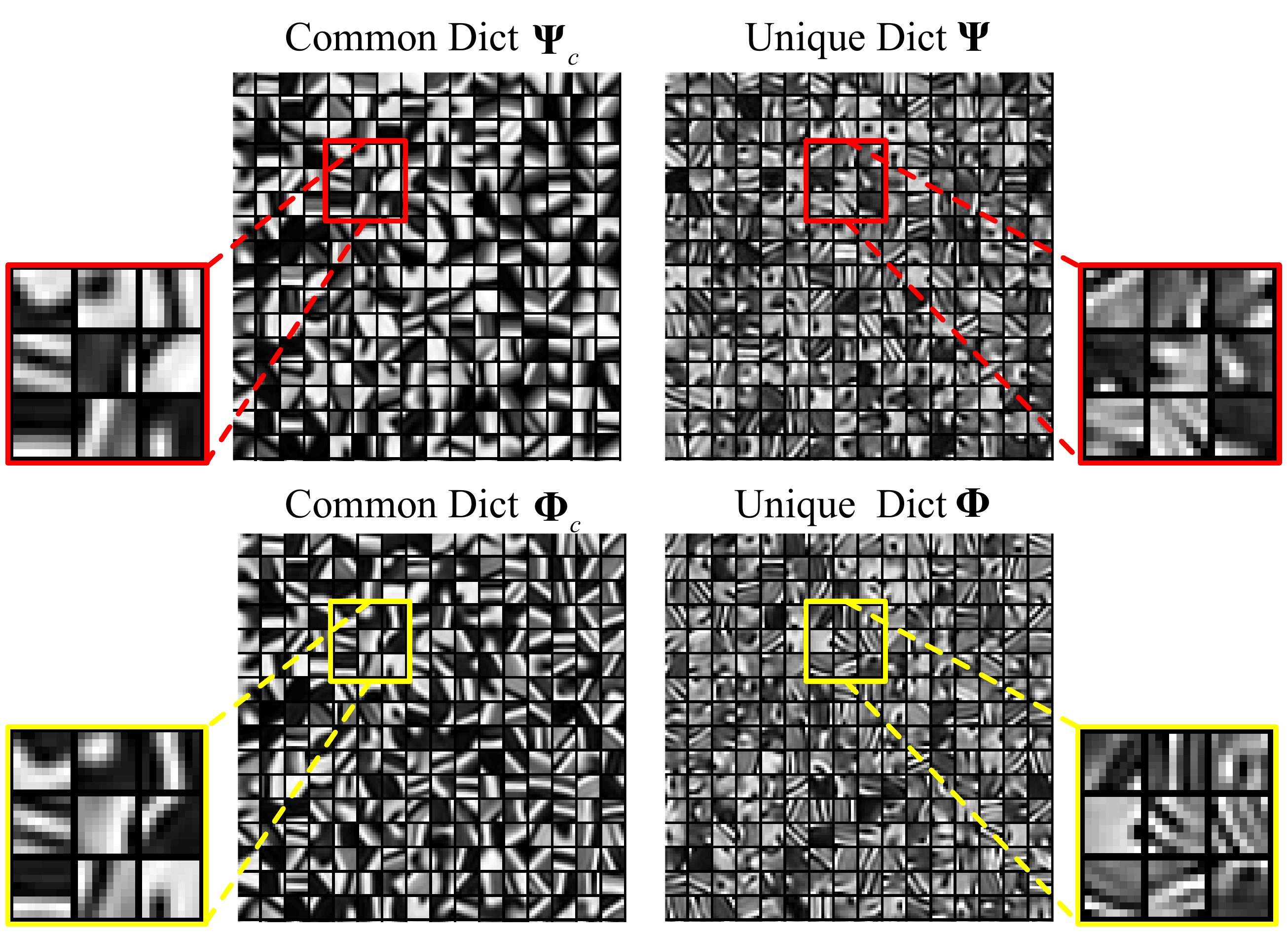}
	\end{minipage} 
	
	\caption{Learned coupled dictionaries from T1 and T2-weighted contrasts; 256 atoms are shown. The top row displays the common and unique dictionaries for the T1-weighted contrast. The bottom row displays dictionaries learned from corresponding T2-weighted contrast. It can be seen that common dictionaries exhibit atoms with similar structure.
	}
	\label{Fig:Dicts}
\end{figure}

Fig.~\ref{Fig:Dicts} shows the learned coupled dictionaries from the T1- and T2- weighted MRI images. It can be seen that the atom pairs from common dictionaries $\boldsymbol{\Psi}_{c}, \boldsymbol{\Phi}_{c}$ capture associated edges, blobs and textures with the same direction and location. Most of them exhibit considerable resemblance to each other, but with opposite intensity. This phenomenon is consistent with MRI characteristics, such as Cerebrospinal fluid (CSF) being dark in T1-weighted contrast and bright in T2-weighted. This outcome indicates that the common dictionaries are able to capture the similarity between T1-weighted and T2-weighted contrasts. In comparison, the learned unique dictionaries $\boldsymbol{\Psi} , \boldsymbol{\Phi}$ represent the disparities of these modalities and therefore rarely exhibit similarity.




\vspace{-0.3cm}

\vspace{-0.1cm}

\section{Conclusion}
\label{sec:Conclusion}

\vspace{-0.3cm}

We presented an adaptive multi-contrast MRI reconstruction framework that capitalizes on both patch-based sparsity priors induced by coupled dictionaries and structure similarity priors from the guidance contrast. The coupled dictionaries are trained directly on the target images and thus are adaptive to the contrast of interest. In addition, they also capture the correlations between T1- and T2-weighted contrasts, thereby beneficial information can be extracted from the guidance contrast to aid the reconstruction of the target contrast. Practical experiments demonstrate the superior performance of our design and significant advantage over competing methods. 
In the future, we will adapt our algorithm for other practical sampling patterns, such as radial sampling and also explore the impact of noise.

\clearpage

%
%



\pagebreak



\clearpage


%




\section*{Appendix A: CDLMRI}

We provide more details in relation to Section \ref{ssec:CDLMRI} CDLMRI in this appendix.


\mypar{Stage 1) Coupled Dictionary Learning (details)}
The dictionary update step is to solve the optimization problem:
\begin{equation} \label{Eq:CDLMRI_DictUpdateStep}
\small
\begin{array}{cl}
\underset{
	\begin{subarray}{c}
	\boldsymbol{\Psi}_c, \boldsymbol{\Psi}, \boldsymbol{\Phi}_c, \boldsymbol{\Phi} \\
	\end{subarray}
}{\text{minimize}}
&  \!\!\!\!		
\sum\limits_{ij} 
\Big\{
\| \mathbf{R}_{ij} \mathbf{x}^{(1)} - ( \boldsymbol{\Psi}_c \mathbf{z}_{ij} + \boldsymbol{\Psi} \mathbf{u}_{ij} ) \|_2^2 
\\
&  \!\!\!\!
\quad \;
+
\| \mathbf{R}_{ij} \mathbf{x}^{(2)} - ( \boldsymbol{\Phi}_c \mathbf{z}_{ij} + \boldsymbol{\Phi} \mathbf{v}_{ij} ) \|_2^2 
\Big\}
\\
\text{subject to}
& \!\!\!\!
\left\| \begin{bmatrix}
\boldsymbol{\psi}_{ck} \\ \boldsymbol{\phi}_{ck} 
\end{bmatrix} \right\|_2^2 \leq 1, 
\|\boldsymbol{\psi}_k \|_2^2 \leq 1, 
\|\boldsymbol{\phi}_k \|_2^2 \leq 1, 
\, \forall k 
\end{array}
\end{equation}
\noindent
Given a subset of the patches to constitute the training dataset $\mathbf{X}^{(1)}=[\cdots, \mathbf{x}_{ij}^{(1)}, \cdots]$ and $\mathbf{X}^{(2)}=[\cdots, \mathbf{x}_{ij}^{(2)}, \cdots]$ in Stage 1), the optimization problem~\eqref{Eq:CDLMRI_DictUpdateStep} is equivalent to:
\begin{equation*} 
\small
\begin{array}{cl}
\underset{
	\begin{subarray}{c}
	\boldsymbol{\Psi}_c, \boldsymbol{\Psi}, \boldsymbol{\Phi}_c, \boldsymbol{\Phi} \\
	\end{subarray}
}{\text{minimize}}
&  \!\!\!\!		
\left\|
\begin{bmatrix} 
\mathbf{X}^{(1)} - \boldsymbol{\Psi} \mathbf{U} \\ 
\mathbf{X}^{(2)} -  \boldsymbol{\Phi} \mathbf{V}
\end{bmatrix} 
- 	
\begin{bmatrix} 
\boldsymbol{\Psi}_c \\ 
\boldsymbol{\Phi}_c
\end{bmatrix} 
\mathbf{Z} \right\|_F^2
\\
\text{subject to}
& \!\!\!\!
\left\| \begin{bmatrix}
\boldsymbol{\psi}_{ck} \\ \boldsymbol{\phi}_{ck} 
\end{bmatrix} \right\|_2^2 \leq 1, 
\|\boldsymbol{\psi}_k \|_2^2 \leq 1, 
\|\boldsymbol{\phi}_k \|_2^2 \leq 1, 
\, \forall k 
\end{array}
\end{equation*}
\noindent
where, $\mathbf{Z}=[\cdots, \mathbf{z}_{ij}^{(1)}, \cdots]$, $\mathbf{U}=[\cdots, \mathbf{u}_{ij}^{(1)}, \cdots]$, $\mathbf{V}=[\cdots, \mathbf{v}_{ij}^{(1)}, \cdots]$. Taking the dictionary update of $\boldsymbol{\Psi}_c$ and $\boldsymbol{\Phi}_c$ for example, we update the atom pairs one by one. For the $k$-th atom pair $\boldsymbol{\psi}_c$ and $\boldsymbol{\phi}_c$, we can immediately establish that
\begin{equation*} 
\small
\begin{array}{rl}
\mathbf{d}_k \leftarrow
\underset{
	\begin{subarray}{c}
	\mathbf{d} \\
	\end{subarray}
}{\text{min}}
&  \!\!\!\!		
\left\|
\begin{bmatrix} 
\mathbf{X}^{(1)} - \boldsymbol{\Psi} \mathbf{U} \\ 
\mathbf{X}^{(2)} -  \boldsymbol{\Phi} \mathbf{V}
\end{bmatrix} 
- 	
\begin{bmatrix} 
\boldsymbol{\Psi}_c \\ 
\boldsymbol{\Phi}_c
\end{bmatrix} 
\mathbf{Z} 
+
\begin{bmatrix}
\boldsymbol{\psi}_{ck}  \\
\boldsymbol{\phi}_{ck}  \\
\end{bmatrix}
\mathbf{z}^k
-
\mathbf{d} \mathbf{z}^k
\right\|_F^2
\\
\text{s.t.}
& \!\!\!\!
\left\| \mathbf{d} \right\|_2^2 \leq 1, 
\end{array}
\end{equation*}
\noindent
By expanding the Frobenius norm and removing the constant term, it turns out that the above problem is equivalent to the optimization problem
\begin{equation*} 
\small
\begin{array}{rl}
\underset{
	\begin{subarray}{c}
	\mathbf{d} \\
	\end{subarray}
}{\text{min}}
&  \!\!\!\!	
\frac{1}{2}
\left\|
\mathbf{d} \mathbf{z}^k
\right\|_F^2	
- \mathbf{d}^T
\left(
\begin{bmatrix} 
\mathbf{X}^{(1)} - \boldsymbol{\Psi} \mathbf{U} \\ 
\mathbf{X}^{(2)} -  \boldsymbol{\Phi} \mathbf{V}
\end{bmatrix} 
- 	
\begin{bmatrix} 
\boldsymbol{\Psi}_c \\ 
\boldsymbol{\Phi}_c
\end{bmatrix} 
\mathbf{Z} 
+
\begin{bmatrix}
\boldsymbol{\psi}_{ck}  \\
\boldsymbol{\phi}_{ck}  \\
\end{bmatrix}
\mathbf{z}^k
\right) 
\mathbf{z}^{k^T}
\\
\text{s.t.}
& \!\!\!\!
\left\| \mathbf{d} \right\|_2^2 \leq 1, 
\end{array}
\end{equation*}
where $\mathbf{z}^k$ denotes the $k$-th row of $\mathbf{Z}$.\footnote{Note that $\mathbf{z}^k$ is a row vector resulting from the derivative w.r.t the $k$-th atom pair, while $\mathbf{z}_{ij}$ is a column vector corresponding to the $ij$-th patch pair.}
\noindent
We compute the derivative of the objective w.r.t. $\mathbf{d}$, leading to a norm equation:
\begin{align*}
&
\mathbf{d}_k  
\leftarrow \frac{1}{\mathbf{z}^k {\mathbf{z}^k}^T}
\left(
\begin{bmatrix} 
\mathbf{X}^{(1)} - \boldsymbol{\Psi} \mathbf{U} \\ 
\mathbf{X}^{(2)} -  \boldsymbol{\Phi} \mathbf{V}
\end{bmatrix} 
- 	
\begin{bmatrix} 
\boldsymbol{\Psi}_c \\ 
\boldsymbol{\Phi}_c
\end{bmatrix} 
\mathbf{Z} \right)
{\mathbf{z}^k}^T + 
\begin{bmatrix}
\boldsymbol{\psi}_{ck}  \\
\boldsymbol{\phi}_{ck}  \\
\end{bmatrix}
\end{align*}
\noindent
Then, we apply the $\ell_2$ norm constraint. 
\begin{align*}
&
\begin{bmatrix}
\boldsymbol{\psi}_{ck}  \\
\boldsymbol{\phi}_{ck}  \\
\end{bmatrix}
= \mathbf{d}_k \leftarrow \frac{\mathbf{d}_k}{\max(\|\mathbf{d}_k \|_2, 1)} 
\end{align*}
The dictionary update of $\boldsymbol{\Psi}$ and $\boldsymbol{\Phi}$ is performed in a similar way. In order to accelerate the training, the proposed algorithm can be updated to online training version without difficulty.

\mypar{Stage 3) $k$-space Consistency Enforcing (details)}
In this stage, we aim to enforce consistency between the denoised image and its measurements in the $k$-space domain. In particular, given the estimated patches $\hat{\mathbf{x}}^{(1)}_{ij}$ from Stage 2), this step is formulated as a least square problem:
\begin{equation} \label{Eq:CDLMRI_ReconUpdate1}
\begin{array}{cl}
\underset{
	\begin{subarray}{c}
	\mathbf{x}^{(1)}
	\end{subarray}
}{\text{min}}
& 
\!\!\!\!			
\sum\limits_{ij} \left\| \mathbf{R}_{ij} \mathbf{x}^{(1)} - \hat{\mathbf{x}}^{(1)}_{ij} \right\|_2^2
+ \nu_1 \left \|\mathbf{F}_{u1} \mathbf{x}^{(1)} - \mathbf{y}^{(1)}  \right \|_2^2
\end{array} \,,
\end{equation}
which admits an analytical solution satisfying the normal equation
\begin{equation} \label{Eq:CDLMRI_ReconUpdate3}
\small
\left( \sum_{ij} \mathbf{R}_{ij}^H \mathbf{R}_{ij} + \nu_1 \mathbf{F}_{u1}^H \mathbf{F}_{u1} \right) \mathbf{x}^{(1)} 
= 
\sum_{ij} \mathbf{R}_{ij}^H \hat{\mathbf{x}}^{(1)}_{ij}
+\nu_1 \mathbf{F}_{u1}^H \mathbf{y}^{(1)} \,,
\end{equation}
\noindent
where the superscript $()^H$ denotes the Hermitian transpose operation. 
The term $ \sum_{ij} \mathbf{R}_{ij}^H \mathbf{R}_{ij} \in \mathcal{C}^{N \times N}$ is a diagonal matrix where each diagonal entry is the number of overlapping patches at the corresponding pixel location in $\mathbf{x}^{(1)}$. Assuming that patches wrap around at image boundaries, the number of overlapping patches at each pixel is the same, denoted by $\beta$.\footnote{In particular, $\beta = n$ when the overlap stride $r = 1$, where the \textit{overlap stride} is defined as the distance in pixels between corresponding pixel locations in adjacent image patches.}
Thus, the term $\frac{1}{\beta} \sum_{ij} \mathbf{R}_{ij}^H \hat{\mathbf{x}}^{(1)}_{ij} $ represents the denoised image $\hat{\mathbf{x}}^{(1)}$, where the intensity value of each pixel is the average of all the overlapping patches that cover this pixel. 
Multiplying by the normalized full Fourier transform matrix $\mathbf{F}$ on the both sides of equation~\eqref{Eq:CDLMRI_ReconUpdate3} leads to
\vspace{-0.2cm}
\begin{multline} \label{Eq:CDLMRI_ReconUpdate5}
\small
\left( \mathbf{F} \sum_{ij} \mathbf{R}_{ij}^H \mathbf{R}_{ij} \mathbf{F}^H + \nu_1 \mathbf{F} \mathbf{F}_{u1}^H \mathbf{F}_{u1} \mathbf{F}^H \right) \mathbf{F} \mathbf{x}^{(1)} 
\\
= 
\mathbf{F} \sum_{ij}  \mathbf{R}_{ij}^H \hat{\mathbf{x}}^{(1)}_{ij}  +\nu_1 \mathbf{F} \mathbf{F}_{u1}^H \mathbf{y}^{(1)} \,.
\end{multline}
\noindent
The matrix $\mathbf{F} \mathbf{F}_{u1}^H \mathbf{F}_{u1} \mathbf{F}^H$ is a diagonal matrix consisting of ones (corresponding to sampling locations in $k$-space) and zeros. Under the "wrap around" assumption, $\mathbf{F} \sum_{ij} \mathbf{R}_{ij}^H \mathbf{R}_{ij} \mathbf{F}^H = \beta \mathbf{I}_P$. Thus, the matrix pre-multiplying $\mathbf{F} \mathbf{x}^{(1)}$ in~\eqref{Eq:CDLMRI_ReconUpdate5} is diagonal and trivially invertible. The vector $\mathbf{F} \mathbf{F}_{u1}^H \mathbf{y}^{(1)}$ represents the zero-filled Fourier measurements. Dividing both sides of \eqref{Eq:CDLMRI_ReconUpdate5} by the constant $\beta$ to obtain
\begin{equation*} 
\tilde{\mathbf{y}}^{(1)}_{pq} = 
\left\{
\begin{aligned}
& \left( \mathbf{F} \hat{\mathbf{x}}^{(1)} \right)_{pq}, 	& \!\!\!\! (p, q) \notin \Omega^{(1)}
\\
&  \frac{ 1 }{1 + \tilde{\nu}_1} \left( \mathbf{F} \hat{\mathbf{x}}^{(1)} + \tilde{\nu}_1 \mathbf{F} \mathbf{F}_{u1}^H \mathbf{y}^{(1)} \right)_{pq}, 	& \!\!\!\! (p, q) \in \Omega^{(1)}
\end{aligned}
\right.
\end{equation*}

\noindent
where $\tilde{\nu}_1 = \nu_1 / \beta$, $\hat{\mathbf{x}}^{(1)}=\frac{1}{\beta} \sum_{ij} \mathbf{R}_{ij}^H \hat{\mathbf{x}}^{(1)}_{ij} $ denotes the denoised image. We denote by $\Omega^{(1)}$ the subset of $k$-space that has been sampled and by $\tilde{\mathbf{y}}^{(1)}_{pq}$ the updated value at location ($p, q$) in the $k$-space. Note that~\eqref{Eq:CDLMRI_ReconUpdate9} uses the dictionaries that were learned in Stage 1) to interpolate the non-sampled Fourier frequencies, and update the sampled frequencies. Then, we immediately obtain the solution:
\begin{equation} \label{Eq:CDLMRI_ReconUpdateEnd}
\tilde{\mathbf{x}}^{(1)}=\mathbf{F}^H \tilde{\mathbf{y}}^{(1)}
\end{equation}
where $\mathbf{F}^H$ denotes the conjugate Fourier transform matrix. $\tilde{\mathbf{y}}^{(1)}$ denotes the estimated $k$-space samples as in \eqref{Eq:CDLMRI_ReconUpdate9}. In other words, the estimation $\tilde{\mathbf{x}}^{(1)}$ is obtained by inverse DFT of $\tilde{\mathbf{y}}^{(1)}$. Then the process returns to the Stage 1). The whole process is shown in Algorithm~\ref{Alg:CDLMRI}.

\section*{Appendix B: More Experiments}
Tissue can be characterized by two different relaxation times -- T1 (longitudinal relaxation time) and T2 (transverse relaxation time).\footnote{T1 (longitudinal relaxation time) is a measure of the time taken for excited spinning protons to realign with the external magnetic field and return to equilibrium. T2 (transverse relaxation time) is a measure of the time taken for excited spinning protons to lose phase coherence among the nuclei spinning perpendicular to the main field.}
T1-weighted and T2-weighted pair of MRI scans are two basic types of multi-contrast data, where the former is produced by using short TE and TR times and conversely the latter is produced by using longer TE (Time to Echo) and TR (Repetition Time) times.
%
In general, T1-weighted MRI images results in highlighted/bright fat tissue, such as subcutaneous fat (SC fat) and bone marrow, and suppressed/dark water-based tissue, such as Cerebrospinal fluid (CSF). In contrast, T2-weighted MRI images highlight both fat tissue and water-based tissue. Therefore, the correlation of T1-weighted and T2-weighted is complex, instead of simple reverse mapping relationship.

\begin{figure*}[t]
	\begin{multicols}{2}  
		\centering
		\begin{minipage}[b]{0.3\linewidth}
			\raggedright
			\includegraphics[height = 2.2cm]{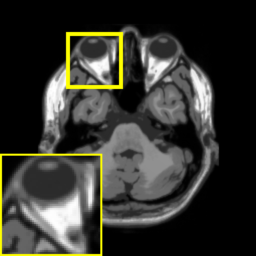}
			\scriptsize Groundtruth T1
		\end{minipage} 
		\begin{minipage}[b]{0.3\linewidth}
			\raggedright
			\includegraphics[height = 2.2cm]{Cart4Fold_CartMask_Fold4.png}
			\scriptsize Mask
		\end{minipage} 
		\begin{minipage}[b]{0.3\linewidth}
			\raggedright
			\includegraphics[height = 2.2cm]{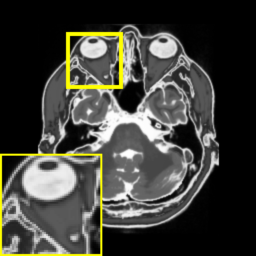}
			\scriptsize Guidance T2
		\end{minipage} 
		\\
		\vspace{0.2cm}
		\begin{minipage}[b]{0.3\linewidth}
			\raggedright
			\includegraphics[height = 2.2cm]{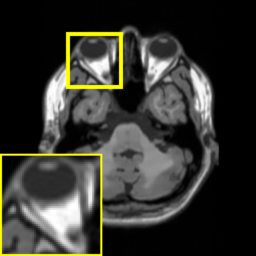}
		\end{minipage} 
		\begin{minipage}[b]{0.3\linewidth}
			\raggedright
			\includegraphics[height = 2.2cm]{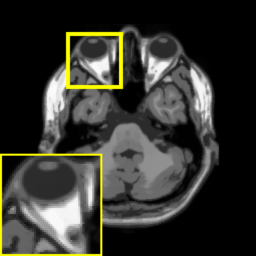}
		\end{minipage} 
		\begin{minipage}[b]{0.3\linewidth}
			\raggedright
			\includegraphics[height = 2.2cm]{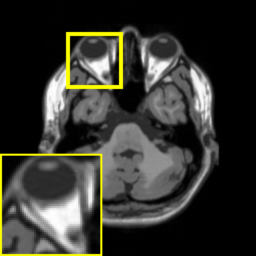}
		\end{minipage} 
		\\
		\begin{minipage}[b]{0.3\linewidth}
			\raggedright
			\includegraphics[height = 2.2cm]{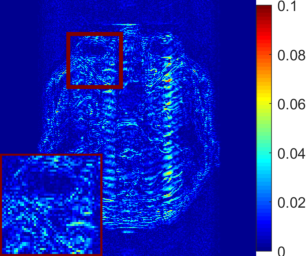}
		\end{minipage} 
		\begin{minipage}[b]{0.3\linewidth}
			\raggedright
			\includegraphics[height = 2.2cm]{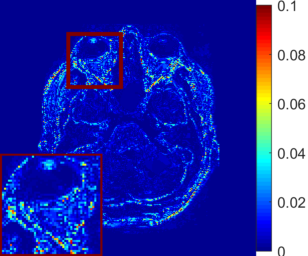}
		\end{minipage} 
		\begin{minipage}[b]{0.3\linewidth}
			\raggedright
			\includegraphics[height = 2.2cm]{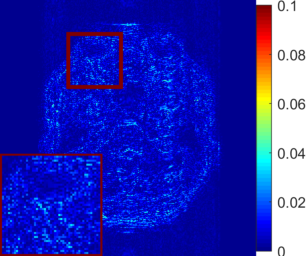}
		\end{minipage} 
		\\
		\vspace{0.2cm}
		\begin{minipage}[b]{0.3\linewidth}
			\raggedright \scriptsize DLMRI 
			\\PSNR = 39.7dB. 
		\end{minipage} 
		\begin{minipage}[b]{0.3\linewidth}
			\raggedright \scriptsize STVMRI 
			\\PSNR = 40.7dB. 
		\end{minipage} 
		\begin{minipage}[b]{0.3\linewidth}
			\raggedright \scriptsize CDLMRI, 
			\\PSNR = 43.3dB. 
		\end{minipage} 
		\\
		\begin{minipage}[t]{\linewidth}
			\centering \footnotesize (a) simulated MRI with 4 fold 1D random under-sampling. 
		\end{minipage} 
		\\
		\begin{minipage}[b]{0.3\linewidth}
			\raggedright
			\includegraphics[height = 2.2cm]{Cart4Fold_ImgNo4_X_true.png}
			\scriptsize Groundtruth T1
		\end{minipage} 
		\begin{minipage}[b]{0.3\linewidth}
			\raggedright
			\includegraphics[height = 2.2cm]{Cart4Fold_CartMask_Fold4.png}
			\scriptsize Mask
		\end{minipage} 
		\begin{minipage}[b]{0.3\linewidth}
			\raggedright
			\includegraphics[height = 2.2cm]{Cart4Fold_ImgNo4_SI_true.png}
			\scriptsize Guidance T2
		\end{minipage} 
		\\
		\vspace{0.2cm}
		\begin{minipage}[b]{0.3\linewidth}
			\raggedright
			\includegraphics[height = 2.2cm]{Cart4Fold_ImgNo4_Fold4_X_DLMRI.png}
		\end{minipage} 
		\begin{minipage}[b]{0.3\linewidth}
			\raggedright
			\includegraphics[height = 2.2cm]{Cart4Fold_ImgNo4_Fold4_X_STVMRI.png}
		\end{minipage} 
		\begin{minipage}[b]{0.3\linewidth}
			\raggedright
			\includegraphics[height = 2.2cm]{Cart4Fold_ImgNo4_Fold4_X_CDLMRI.png}
		\end{minipage} 
		\\
		\begin{minipage}[b]{0.3\linewidth}
			\raggedright
			\includegraphics[height = 2.2cm]{Cart4Fold_ImgNo4_Fold4_X_resid_DLMRI.png}
		\end{minipage} 
		\begin{minipage}[b]{0.3\linewidth}
			\raggedright
			\includegraphics[height = 2.2cm]{Cart4Fold_ImgNo4_Fold4_X_resid_STVMRI.png}
		\end{minipage} 
		\begin{minipage}[b]{0.3\linewidth}
			\raggedright
			\includegraphics[height = 2.2cm]{Cart4Fold_ImgNo4_Fold4_X_resid_CDLMRI.png}
		\end{minipage} 
		\\
		\vspace{0.2cm}
		\begin{minipage}[b]{0.3\linewidth}
			\raggedright \scriptsize DLMRI 
			\\PSNR = 33.4dB. 
		\end{minipage} 
		\begin{minipage}[b]{0.3\linewidth}
			\raggedright \scriptsize STVMRI 
			\\PSNR = 33.7dB. 
		\end{minipage} 
		\begin{minipage}[b]{0.3\linewidth}
			\raggedright \scriptsize CDLMRI 
			\\PSNR = 36.1dB. 
		\end{minipage} 
		\\
		\begin{minipage}[t]{\linewidth}
			\centering \footnotesize (b) clinical MRI with 4 fold 1D random under-sampling. 
		\end{minipage} 
		\\
	\end{multicols}	
	
	\vspace{-0.5cm}
	
	\caption{Reconstruction for T1-weighted MRI, with fully-sampled T2-weighted version as reference using 4 fold Cartesian 1D random under-sampling. The first row shows the groundtruth T1-weighted contrast, sampling mask and guidance T2-weighted contrast. The second and third rows show the reconstructed images and the corresponding residual error from DLMRI~\cite{ravishankar2011mr}, STVMRI~\cite{ehrhardt2016multicontrast}, and the proposed CDLMRI. It can be seen that the proposed approach reliably reconstructs fine details and substantially suppresses aliasing, noise and artifacts, leading to the smallest residual error.}
	\label{Fig:TestIm4_1D_Appen}
\end{figure*}

\begin{figure*}[th]
	\begin{multicols}{2}  
		\centering
		\begin{minipage}[b]{0.3\linewidth}
			\raggedright
			\includegraphics[height = 2.2cm]{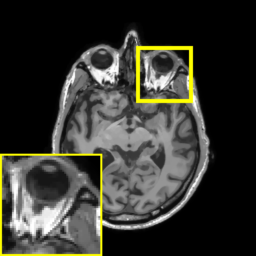}
			\scriptsize Groundtruth T1
		\end{minipage} 
		\begin{minipage}[b]{0.3\linewidth}
			\raggedright
			\includegraphics[height = 2.2cm]{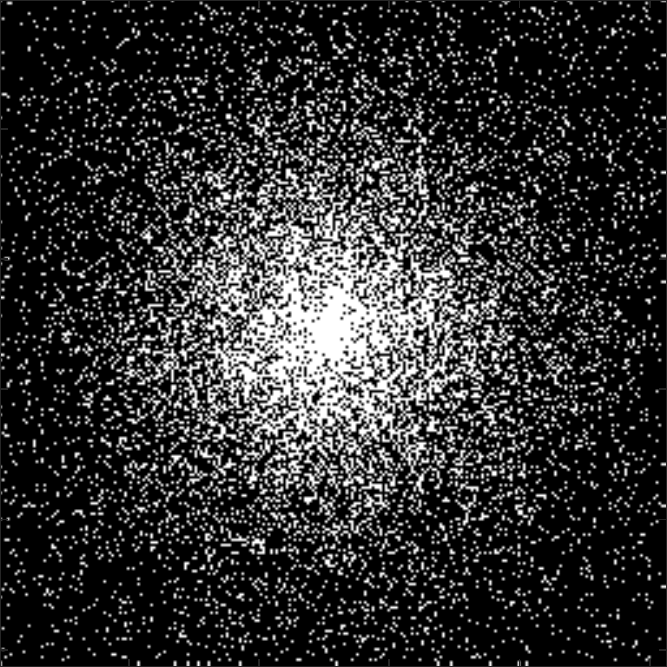}
			\scriptsize Mask
		\end{minipage} 
		\begin{minipage}[b]{0.3\linewidth}
			\raggedright
			\includegraphics[height = 2.2cm]{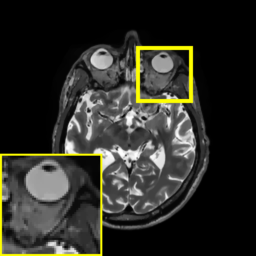}
			\scriptsize Guidance T2
		\end{minipage} 
		\\
		\vspace{0.2cm}
		\begin{minipage}[b]{0.3\linewidth}
			\raggedright
			\includegraphics[height = 2.2cm]{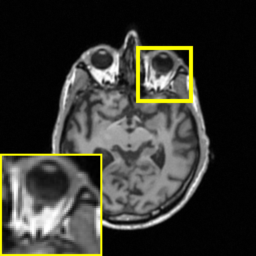}
		\end{minipage} 
		\begin{minipage}[b]{0.3\linewidth}
			\raggedright
			\includegraphics[height = 2.2cm]{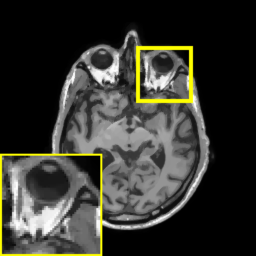}
		\end{minipage} 
		\begin{minipage}[b]{0.3\linewidth}
			\raggedright
			\includegraphics[height = 2.2cm]{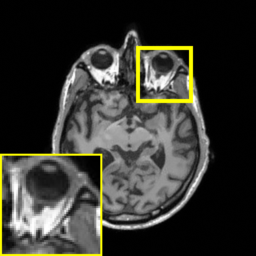}
		\end{minipage} 
		\\
		\begin{minipage}[b]{0.3\linewidth}
			\raggedright
			\includegraphics[height = 2.2cm]{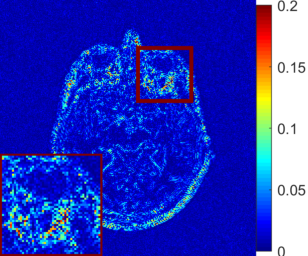}
		\end{minipage} 
		\begin{minipage}[b]{0.3\linewidth}
			\raggedright
			\includegraphics[height = 2.2cm]{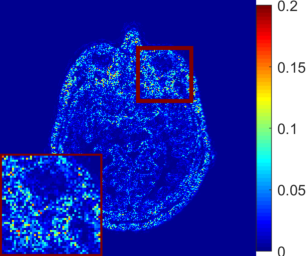}
		\end{minipage} 
		\begin{minipage}[b]{0.3\linewidth}
			\raggedright
			\includegraphics[height = 2.2cm]{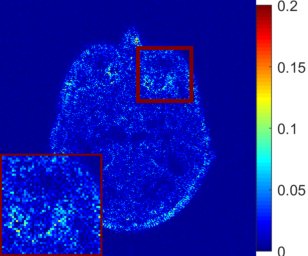}
		\end{minipage} 
		\\
		\vspace{0.2cm}
		\begin{minipage}[b]{0.3\linewidth}
			\raggedright \scriptsize DLMRI 
			\\PSNR = 32.9dB. 
		\end{minipage} 
		\begin{minipage}[b]{0.3\linewidth}
			\raggedright \scriptsize STVMRI 
			\\PSNR = 34.2dB. 
		\end{minipage} 
		\begin{minipage}[b]{0.3\linewidth}
			\raggedright \scriptsize CDLMRI
			\\PSNR = 36.8dB. 
		\end{minipage} 
		\\
		\begin{minipage}[t]{\linewidth}
			\centering \footnotesize (a) 5 fold 2D random under-sampling. 
		\end{minipage} 
		\\
		\begin{minipage}[b]{0.3\linewidth}
			\raggedright
			\includegraphics[height = 2.2cm]{2Drandom_ImgNo4_X_true.png}
			\scriptsize Groundtruth T1
		\end{minipage} 
		\begin{minipage}[b]{0.3\linewidth}
			\raggedright
			\includegraphics[height = 2.2cm]{2Drandom_2DrandomMask_Fold20.png}
			\scriptsize Mask
		\end{minipage} 
		\begin{minipage}[b]{0.3\linewidth}
			\raggedright
			\includegraphics[height = 2.2cm]{2Drandom_ImgNo4_SI_true.png}
			\scriptsize Guidance T2
		\end{minipage} 
		\\
		\vspace{0.2cm}
		\begin{minipage}[b]{0.3\linewidth}
			\raggedright
			\includegraphics[height = 2.2cm]{2Drandom_ImgNo4_Fold20_X_DLMRI.png}
		\end{minipage} 
		\begin{minipage}[b]{0.3\linewidth}
			\raggedright
			\includegraphics[height = 2.2cm]{2Drandom_ImgNo4_Fold20_X_STVMRI.png}
		\end{minipage} 
		\begin{minipage}[b]{0.3\linewidth}
			\raggedright
			\includegraphics[height = 2.2cm]{2Drandom_ImgNo4_Fold20_X_CDLMRI.png}
		\end{minipage} 
		\\
		\begin{minipage}[b]{0.3\linewidth}
			\raggedright
			\includegraphics[height = 2.2cm]{2Drandom_ImgNo4_Fold20_X_resid_DLMRI.png}
		\end{minipage} 
		\begin{minipage}[b]{0.3\linewidth}
			\raggedright
			\includegraphics[height = 2.2cm]{2Drandom_ImgNo4_Fold20_X_resid_STVMRI.png}
		\end{minipage} 
		\begin{minipage}[b]{0.3\linewidth}
			\raggedright
			\includegraphics[height = 2.2cm]{2Drandom_ImgNo4_Fold20_X_resid_CDLMRI.png}
		\end{minipage} 
		\\
		\vspace{0.2cm}
		\begin{minipage}[b]{0.3\linewidth}
			\raggedright \scriptsize DLMRI 
			\\PSNR = 28.5dB. 
		\end{minipage} 
		\begin{minipage}[b]{0.3\linewidth}
			\raggedright \scriptsize STVMRI 
			\\PSNR = 29.0dB. 
		\end{minipage} 
		\begin{minipage}[b]{0.3\linewidth}
			\raggedright \scriptsize CDLMRI, \\PSNR = 30.2dB. 
		\end{minipage} 
		\\
		\begin{minipage}[t]{\linewidth}
			\centering \footnotesize (b) 20 fold 2D random under-sampling. 
		\end{minipage} 
		\\
	\end{multicols}	
	
	\vspace{-0.5cm}
	
	\caption{Reconstruction for T1-weighted MRI, with fully-sampled T2-weighted version as reference using 5 fold and 20 fold 2D random under-sampling, using DLMRI~\cite{ravishankar2011mr}, STVMRI~\cite{ehrhardt2016multicontrast}, and the proposed CDLMRI. The first row shows the groundtruth T1-weighted, sampling mask and guidance modality T2-weighted. The second and third rows show the reconstructed images and the corresponding residual error from DLMRI~\cite{ravishankar2011mr}, STVMRI~\cite{ehrhardt2016multicontrast}, and the proposed CDLMRI. It can be seen that the proposed approach outperform the competing approaches, leading to the smallest residual error.}
	\label{Fig:TestIm4_2D_Appen}
\end{figure*}

In this experiment, we use under-sampled T1-weighted MRI as the target contrast and corresponding fully-sampled T2-weighted as the guidance contrast to replicate the same scenario as in~\cite{ehrhardt2016multicontrast}. Similar to previous approaches\cite{ma2008efficient,yang2010fast,trzasko2009highly,ravishankar2011mr,ehrhardt2016multicontrast,weizman2016reference}, the data acquisition was simulated by retrospectively under-sampling the 2D discrete Fourier transform of clinical magnitude MR images.\footnote{After using the Fourier transform to transform measured k-space data into image space, the image data is of complex type, which is then manipulated for different clinical utility. In clinical practice, magnitude images are nearly exclusively used for diagnosis as it maximizes the signal-to-noise ratio (SNR). Phase-images are occasionally generated in clinical MRI for the depiction of flow and characterization of susceptibility-induced distortions. Therefore, from the perspective of diagnosis, we focus on the magnitude images}. The sampling masks include Cartesian 1D and 2D random sampling. We compare the proposed approach with DLMRI~\cite{ravishankar2011mr} to show the benefits of integrating guidance information into the MRI reconstruction task. 
We also compare with SVTMRI~\cite{ehrhardt2016multicontrast} which uses the structure-guided total variation to integrate the guidance contrast to aid the reconstruction of the target one.Figure~\ref{Fig:TestIm4_1D_Appen} and \ref{Fig:TestIm4_2D_Appen} show reconstruction results for the scenario where a variable density Cartesian mask is employed for under-sampling on the target T1-weighted contrast, with a fully sampled T2-weighted MRI for guidance contrast.

\end{document}